\documentclass{article}
\usepackage{ijcai23}

\usepackage{times}
\usepackage{soul}
\usepackage{url}
\usepackage[hidelinks]{hyperref}
\usepackage[utf8]{inputenc}
\usepackage[small]{caption}
\usepackage{graphicx}
\usepackage{amsmath}
\usepackage{amsthm}
\usepackage{booktabs}
\usepackage{algorithm}
\usepackage{algorithmic}
\usepackage{amsfonts}
\usepackage{latexsym}
\usepackage[switch]{lineno}
\usepackage{multirow}
\usepackage{subcaption}
\usepackage[table]{xcolor}

\definecolor{mygray}{gray}{.9}
\definecolor{sota_blue}{HTML}{0071bc}

\makeatletter
\newcommand{\thickhline}{%
	\noalign {\ifnum 0=`}\fi \hrule height 1pt
	\futurelet \reserved@a \@xhline
}
\makeatother

\title{Discovering Sounding Objects by Audio Queries for Audio Visual Segmentation}

\author{
Shaofei Huang$^{1,2}$
\and
Han Li$^{3}$\and
Yuqing Wang$^4$\and
Hongji Zhu$^4$\and \\
Jiao Dai$^{1,2}$\thanks{Corresponding author}\and
Jizhong Han$^{1,2}$\and
Wenge Rong$^3$\And
Si Liu$^{5,6}$\\
\affiliations
$^1$Institute of Information Engineering, Chinese Academy of Sciences \\
$^2$School of Cyber Security, University of Chinese Academy of Sciences \\
$^3$School of Computer Science and Engineering, Beihang University \\
$^4$Alibaba Group \\
$^5$Institute of Artificial Intelligence, Beihang University \\
$^6$Hangzhou Innovation Institute, Beihang University \\
}

\begin{document}

\maketitle

\begin{abstract}
Audio visual segmentation (AVS) aims to segment the sounding objects for each frame of a given video. 
To distinguish the sounding objects from silent ones, both audio-visual semantic correspondence and temporal interaction are required.
The previous method applies multi-frame cross-modal attention to conduct pixel-level interactions between audio features and visual features of multiple frames simultaneously, which is both redundant and implicit.
In this paper, we propose an Audio-Queried Transformer architecture, AQFormer, where we define a set of object queries conditioned on audio information and associate each of them to particular sounding objects.
Explicit object-level semantic correspondence between audio and visual modalities is established by gathering object information from visual features with predefined audio queries.
Besides, an Audio-Bridged Temporal Interaction module is proposed to exchange sounding object-relevant information among multiple frames with the bridge of audio features. 
Extensive experiments are conducted on two AVS benchmarks to show that our method achieves state-of-the-art performances, especially $7.1\%$ $M_{\mathcal{J}}$ and $7.6\%$ $M_{\mathcal{F}}$ gains on the MS3 setting.
\end{abstract}

\section{Introduction}
It is intuitive that objects are characterized not only by what they look like but also by the sounds they make.
By excavating the correspondence between visual and acoustic information of objects, it is helpful to facilitate the understanding of their characteristics.
In this work, we focus on the problem of audio visual segmentation (AVS)~\cite{AVS}, which aims to identify and segment the sounding objects in each frame of the given video.
Unlike audio-visual correspondence~\cite{L3Net} or sound source localization~\cite{arandjelovic2018objects} that only produce the interval of audible frames or the heatmap centered on the sounding objects, AVS requires more fine-grained correspondence between audio and visual modalities.
In addition, the situation of sounding in each video clip is dynamically changing over time.
For example, different objects may make sounds at different time periods, and multiple objects may make sounds at the same time period.
The uncertainty of target objects also makes AVS even more challenging than other video object segmentation tasks~\cite{VOS,URVOS} where the target object remains the same for the whole video.
\begin{figure}[!t]
    \centering
    \includegraphics[width=\linewidth]{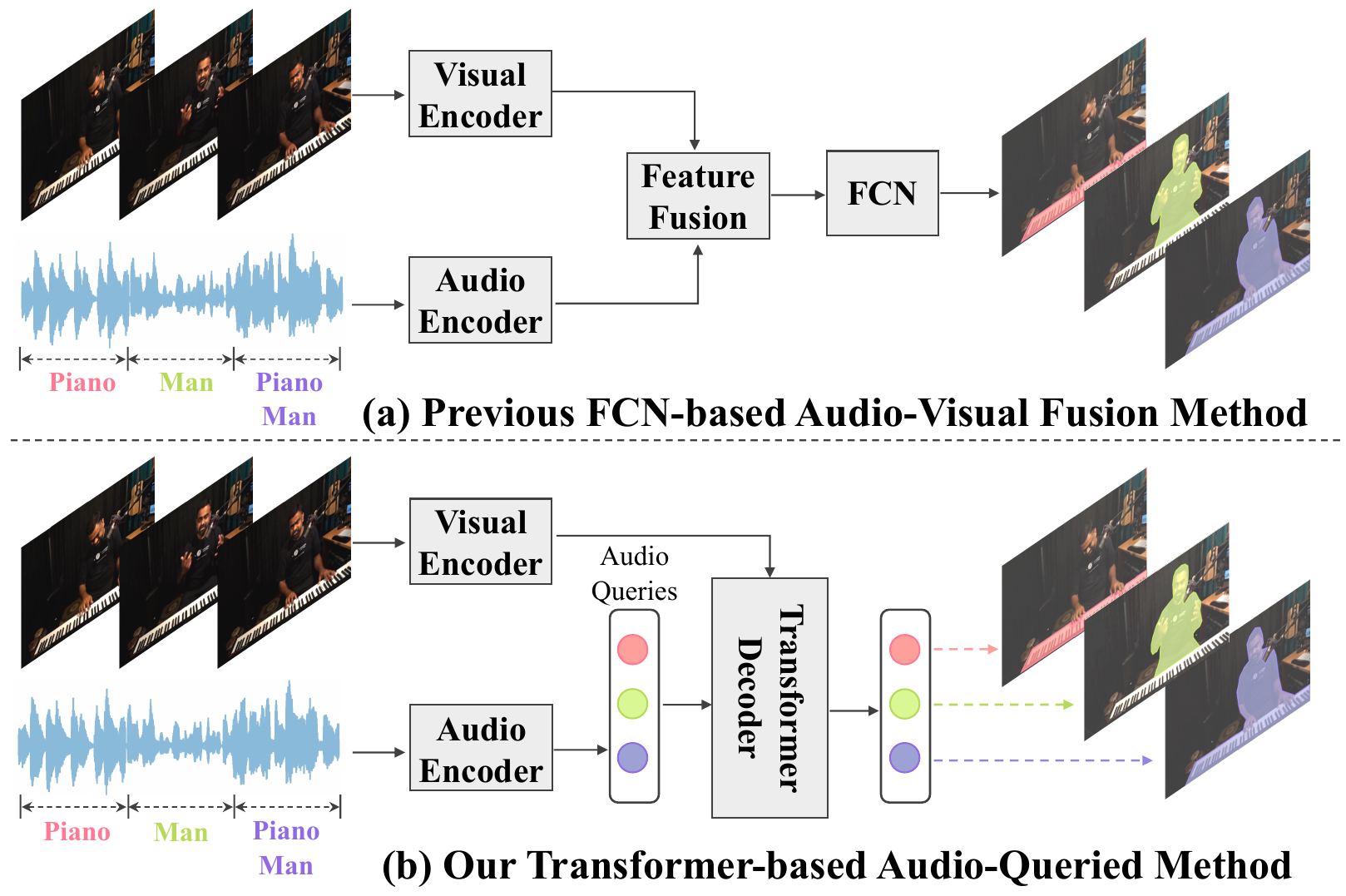}
    \caption{Comparison between different methods. (a) The previous method establishes pixel-level audio-visual correspondence by fusing audio and visual features with cross-modal attention. (b) Our method establishes object-level audio-visual correspondence using audio queries to gather object information from visual features.}
    \label{fig:intro}
\end{figure}

There exist two challenges in identifying the sounding objects under complex audio-visual scenes.
On the one hand, objects of different categories typically differ from each other in the sounds they emit, indicating that we can distinguish them by their acoustic characteristics.
As illustrated in Figure~\ref{fig:intro}, there is an obvious difference between the waveform diagrams of the human voice and the guitar sound.
Thus, it is critical to establish proper semantic correspondence between audio and visual features.
On the other hand, due to the similarity of acoustic characteristics between objects of the same category, it is difficult to distinguish them with their audio features, which may lead to confusion between them.
For example, there are two men in the video of Figure~\ref{fig:intro} who sing in the first frame and second frame respectively, making it hard to tell them apart without additional clues.
Therefore, temporal contexts such as action and mouth movement can be exploited to locate the sounding objects in such cases.

As shown in Figure~\ref{fig:intro}(a), the previous method~\cite{AVS} first applies a multi-frame cross-modal attention module to fuse the visual features and the audio features extracted by a visual encoder and an audio encoder respectively, and then makes predictions on the fused multi-modal features with an FCN head.
It has some limitations in dealing with the above two challenges.
First, the cross-modal attention interacts between visual and audio features only at the pixel level, which is too implicit and may include unwanted background pixels in the images. 
Such pixel-level audio-visual semantic correspondence leads to an insufficient understanding of the sounding characteristics of different objects.
Second, the temporal information modeling is realized in a mixed manner by the attention module, where each pixel of the video frames densely interacts with all pixels through the attention mechanism.
This way of dense temporal interaction tends to be inefficient and redundant in capturing the temporal movements of sounding objects.

In this work, we propose a new Audio-Queried Transformer architecture, \textbf{AQFormer}, to tackle the above two challenges.
\textit{For the first one}, we define a set of audio-conditioned object queries to build explicit object-level correspondence between audio and visual modalities.
As illustrated in Figure~\ref{fig:intro}(b), each audio query is generated by the audio feature of one frame and is associated with one or more objects that make sounds in the corresponding frame but may also exist in other frames.
By gathering object information from the visual features, the audio queries gradually encode the global information of the associated sounding objects, so that the generated object embeddings can be used to produce masks for the corresponding frames.
An auxiliary loss is also designed to constrain the object-level feature similarity between the audio queries and visual features, which helps to learn more discriminative object representations.
\textit{For the second one}, in order to model the temporal pattern of the sounding objects, we enable each audio query to gather object information from the visual features of all the video frames.
In this way, the temporal information is integrated into the object embeddings to facilitate the distinguishment between visually confusing objects.
In addition, to further endow the visual features with sounding object-aware temporal context, we propose an Audio-Bridged Temporal Interaction (ABTI) module which leverages 
the audio feature as a medium to bridge the interaction between different frames.
Concretely, the audio features are first utilized to gather the audio-relevant visual features of each frame.
The compact features of different frames interact with each other to exchange information in a denoised manner and are then remapped back to the corresponding visual feature to enhance it with temporal context.
The ABTI module is densely inserted after each stage of the visual encoder to generate temporal-aware visual features.

Our contributions are summarized as follows:

(1) We propose a novel Audio-Queried Transformer Architecture, AQFormer, for audio visual segmentation. 
A set of object queries conditioned on audio information are defined and utilized to gather visual information of their associated sounding objects for mask prediction.

(2) We design an Audio-Bridged Temporal Interaction (ABTI) module to exchange sounding object-relevant information among multiple frames bridged by audio features.

(3) Extensive experiments show that our AQFormer significantly outperforms the previous FCN-based method and achieves state-of-the-art performances on both single-source and multi-source AVS benchmarks.

\section{Related Works}
\subsection{Video Segmentation}
\textbf{Video Semantic Segmentation}.
Generally, the popular convolutional-based segmentation methods~\cite{FCN} make pixel-wise classification to classify each pixel into a predefined category by a fully convolutional layer.
Later works make efforts to capture multi-scale context information~\cite{ASPP} or aggregate features with long-range dependencies~\cite{DANet,OCNet}.
With the development of vision transformer~\cite{VIT,Segmenter} and object query methods~\cite{DETR}, MaskFormer~\cite{MaskFormer} first formulates the segmentation problem as mask classification and defines mask queries to encode region features for mask prediction.
Mask2Former~\cite{Mask2Former} further limits the query scopes to local regions by proposing masked attention.
When extended to video semantic segmentation, most methods focus on the problems of temporal information extraction and inter-frame relationship modeling~\cite{STGRU,DFF}.
STFCN~\cite{STFCN} introduces LSTM~\cite{LSTM} to integrate features from multiple frames sequentially.
Similarly, STGRU~\cite{STGRU} inserts GRU~\cite{GRU} in the pipeline to control the feature fusion at different timesteps.

\textbf{Video Object Segmentation}. 
Video object segmentation (VOS) aims to track and segment foreground objects of a video in a class-agnostic manner.
Generally, additional information is provided as a hint to indicate the target object to segment in VOS tasks~\cite{yang2021associating,yang2022decoupling}.
For example, the object mask of the first frame, i.e., the reference frame, is provided in semi-supervised video object segmentation.
To propagate object masks from the reference frame, STM~\cite{STM} introduces a space-time memory network that collects information from both the reference frame and the previous frame.
CFBI~\cite{CFBI} extracts features from both foreground objects and background regions and matches them with the first frame to refine object masks.
Another conditional VOS task, Referring Video Object Segmentation (RVOS)~\cite{URVOS} provides the linguistic expression that describes the object's characteristics to indicate the target object. 
LBDT~\cite{LBDT} proposes a Language-Bridged Duplex Transfer (LBDT) module which utilizes language as an intermediary bridge to exchange information between spatial and temporal branches.
ReferFormer~\cite{ReferFormer} and MTTR~\cite{MTTR} both adopt multimodal transformer architectures that use text embeddings to restrict the object queries and predict object sequence by linking objects from different frames.

\begin{figure*}[!t]
    \centering
    \includegraphics[width=\linewidth]{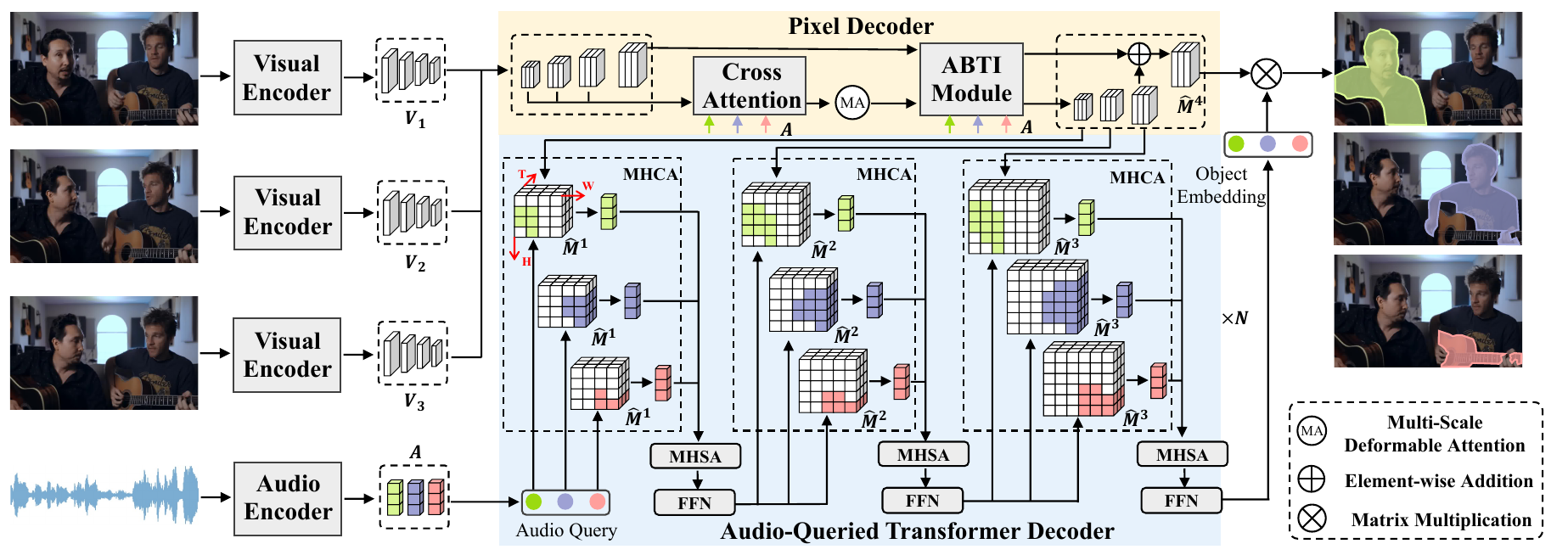}
    \caption{The overall architecture of AQFormer. Given a video clip and its corresponding audio signal, after extracting visual features and audio features, a pixel decoder is followed to enhance the multi-scale visual features with both audio information and temporal information. In the audio-queried transformer decoder, audio queries adaptively gather object information from multi-scale visual features through stacked transformer layers, and the generated object embeddings are then used to produce masks of sounding objects by matrix multiplication. It is worth noting that in each MHCA module, the three cubes all represent the same visual feature and we repeat it for explanation convenience.}
    \label{fig:pipeline}
\end{figure*}

\subsection{Audio-Visual Learning}
Audio and visual signals usually co-occur in the same video, which provides rich semantic information for effective multi-modal learning and scene perception.
Some works~\cite{L3Net,Wav2CLIP,XDC} leverage the natural correspondence between audio and visual information as supervision to learn discriminative multi-modal feature representations.
These methods are generally utilized for pretraining and only segment-level audio-visual correspondence is constructed.
Different from the above methods, Sound Source Localization (SSL)~\cite{arandjelovic2018objects,zhao2019sound,rouditchenko2019self} leverages the relevance between audio and vision regions to distinguish different sound sources and locate audio-related sounding areas.
For example, DMC~\cite{DMC} clusters audio and visual representations within each modality and associates the centroids of both modalities with contrastive learning.
A two-stage framework is proposed in \cite{qian2020multiple} to align features of audio-visual modalities from coarse to fine.
SSL only learns patch-level audio-visual correspondence by producing a heat map of the sounding objects, which is insufficient for applications that require more fine-grained visual results.

In this work, we focus on the problem of audio-visual segmentation (AVS)~\cite{AVS}, which further requires to delineate the shape of the sounding objects at each frame.
A baseline method TPAVI is also proposed in \cite{AVS} to fuse audio features with video features by temporal pixel-wise audio-visual attention mechanism.
Compared with SSL, AVS achieves pixel-level alignment between visual and audio modalities, enabling more comprehensive scene understanding.

\section{Method}
Given a video clip with $T$ frames which is represented by its corresponding visual and audio components, our goal is to obtain the binary mask of sounding objects for each frame.
The overall architecture of our proposed AQFormer is illustrated in Figure~\ref{fig:pipeline}.
A visual encoder is first adopted to extract multi-scale visual features for each frame.
An audio encoder is used to process the input audio signal and obtain the corresponding audio feature.
Then, we feed both the multi-scale visual features and the audio feature into the pixel decoder to obtain the audio-relevant and temporally-enhanced visual features.
Afterward, the first three scales of enhanced visual features are fed into the audio-queried transformer decoder, in which the audio feature is used to generate $T$ conditional object queries to collect object features from the enhanced features, and object embeddings are produced for each frame.
The final predictions are obtained via matrix multiplication between the object embeddings and the largest visual feature output by the pixel encoder.

\subsection{Visual and Audio Encoders}
Given a video clip with $T$ frames, we adopt ResNet-50~\cite{resnet} or PVT-v2~\cite{pvt} as the visual encoder to extract the visual features for each frame separately following~\cite{AVS}.
To capture multi-scale visual context, we extract visual features of $4$ different scales, including $1/32$, $1/16$, $1/8$, and $1/4$ of the original image size.
The visual feature of the $i$-th scale is denoted as $\mathbf{V}^i \in \mathbb{R}^{T\times H^i\times W^i\times C_v^i}$, where $H^i, W^i$ denote the height and width of the $i$-th feature and $C_v^i$ denotes its channel number.
As for the audio input, we first process it with the Short-time Fourier Transform to generate its spectrogram. 
A convolutional network, such as VGGish~\cite{vggish} is then applied to the spectrogram to extract the audio feature $\mathbf{A} \in \mathbb{R}^{T\times C_a}$, where $C_a$ denotes the channel number of $\mathbf{A}$.
In the following sections, we leverage subscript $t$ to index audio or visual feature of the $t$-th frame, e.g., $\mathbf{A}_t\in \mathbb{R}^{C_a}$ and $\mathbf{V}_t^i \in \mathbb{R}^{H^i\times W^i\times C_v^i}$.

\subsection{Pixel Decoder}
In pixel decoder, the multi-scale visual features are enhanced to be aware of both the audio information and temporal context.
First, the audio feature is integrated to enhance the corresponding frame of visual feature by cross-attention.
Since operations performed on each scale of visual features are the same, we take the $i$-th scale as an example and omit the superscript $i$ for simplicity.
The cross-attention is calculated as follows:

\vspace{-0.45cm}
\begin{gather}
\bar{\mathbf{V}}_t = {\rm Softmax}(\frac{f_q(\mathbf{V}_t)\otimes f_k(\mathbf{A}_t)^T}{\sqrt{C_m}})\otimes f_v(\mathbf{A}_t), \\
\mathbf{M}_t = \bar{\mathbf{V}}_t+f_w(\mathbf{V}_t),            
\end{gather}
where $\mathbf{A}_t$ and $\mathbf{V}_t$ are first reshaped to $\mathbb{R}^{1\times C_a}$ and $\mathbb{R}^{HW\times C_v}$ respectively, $\otimes$ denotes matrix multiplication, and $f_q(\cdot)$, $f_k(\cdot)$, $f_v(\cdot)$ and $f_w(\cdot)$ denote linear transform operations to change the channel numbers of input features to $C_m$, $C_m$, $C_f$ and $C_f$ respectively.
Following~\cite{Mask2Former}, the enhanced visual feature $\mathbf{M}^i$ is then reshaped to $\mathbb{R}^{T\times H^i\times W^i\times C_f}$ and fed into the multi-scale deformable attention~\cite{DefDETR} module to obtain the aggregated feature $\bar{\mathbf{M}}^i$.
In order to reduce the computational cost, the above operations are performed on all scales of visual features except for the largest one.
Since the above operations are performed on each frame independently without introducing temporal information, we further propose the Audio Bridged Temporal Interaction (ABTI) module to bridge the cross-frame interaction by audio feature.
As shown in Figure~\ref{fig:pipeline}, ABTI is applied to all scales of visual features, including the largest one $\mathbf{V}^4$ and other aggregated features $\bar{\mathbf{M}}^i, i\in [1,3]$.
The output features of pixel encoder are represented by $\hat{\mathbf{M}}^i \in \mathbb{R}^{T\times H^i \times W^i \times C_f}, i\in [1, 4]$.
The details of ABTI are elaborated in Section~\ref{sec:abti}.

\subsection{Audio-Queried Transformer Decoder}
In the audio-queried transformer decoder, we first define an object query for each frame to refer to the sounding objects in the current frame.
For example, in Figure~\ref{fig:pipeline}, the query denoted by a green circle is associated with the man on the left over the whole video clip, even if this man only makes sounds in the first frame.
Each object query is conditioned on the audio feature of the corresponding frame and is called \textit{audio query} in this paper.
For each video clip, there are $T$ audio queries in total.
To gather information of the associated objects for each audio query, we follow the implementation practice of transformer decoder~\cite{MaskFormer} and revise the cross-attention to enable each audio query to attend to the visual feature of all frames.
All scales of visual features except for the largest one are exploited in the decoder, and audio queries interact with one of these features sequentially from the lowest to the highest resolutions within each stage.
Thus, there are $3$ layers in each stage of the transformer decoder, which are repeated by $N$ times in a round-robin fashion.
We take the $l$-th layer as an example to illustrate how the query features are updated in each transformer layer.
Given the query feature $\mathbf{A}^{l-1}\in \mathbb{R}^{T\times C_f}$ from the last layer and the corresponding reshaped visual feature $\hat{\mathbf{M}}^l\in \mathbb{R}^{TH^lW^l\times C_f}$, query feature $\mathbf{A}^l$ of the $l$-th layer are obtained as follows:

\vspace{-0.35cm}
\begin{gather}
\mathbf{X}^l = {\rm LN}({\rm MHCA}(\mathbf{A}^{l-1}, \hat{\mathbf{M}}^l)) + \mathbf{A}^{l-1}, \\
\bar{\mathbf{X}}^l = {\rm LN}({\rm MHSA}(\mathbf{X}^l)) + \mathbf{X}^l, \\
\mathbf{A}^l = {\rm LN}({\rm FFN}(\bar{\mathbf{X}}^l)) + \bar{\mathbf{X}}^l,
\end{gather}
where we use ${\rm MHCA}(\cdot)$, ${\rm MHSA}(\cdot)$, ${\rm LN}(\cdot)$, and ${\rm FFN}(\cdot)$ to denote multi-head cross-attention, multi-head self-attention, layer normalization, and feedforward network respectively.
The original audio queries $\mathbf{A}^0$ are initialized with the audio feature $\mathbf{A}$ transformed with a linear layer.

Through the above operations, object embeddings $\mathbf{A}^L \in \mathbb{R}^{T\times C_f}$ that outputted by the last decoder layer encode the global information of their associated sounding objects and can be further leveraged to generate the corresponding mask prediction for each frame.
We take the largest visual feature $\hat{\mathbf{M}}_t^4$ as the mask feature and reshape it to $\mathbb{R}^{H^4W^4\times C_f}$.
$\mathbf{A}^L_t$ is also reshaped to the shape of $\mathbb{R}^{1\times C_f}$.
The binary mask of the sounding objects for the $t$-th frame is calculated as follows:
\begin{equation}
\hat{\mathbf{Y}}_t = \sigma(\hat{\mathbf{M}}_t^4 \otimes (\mathbf{A}^L_t)^T),
\end{equation}
where $\sigma(\cdot)$ denotes sigmoid operation. 
$\hat{\mathbf{Y}}_t$ is then reshaped and resized to match the size of the original images. 

To further facilitate the learning of discriminative object-level features, we also propose an auxiliary loss, i.e., $\mathcal{L}_{sim}$, to constrain the object-level feature similarity between $\mathbf{A}^l$ and $\hat{\mathbf{M}}^l$.
The auxiliary loss is formatted as follows:
\begin{equation}
\label{eq:simloss}
\mathcal{L}_{sim} = \sum_{l=1}^{L}{\sum_{p=1}^{T}{\sum_{q=1}^{T}{\vert {\rm Sim}(\mathbf{A}^l_p, \mathbf{A}^l_q)- {\rm Sim}(\mathbf{Z}^l_p, \mathbf{Z}^l_q)\vert}}},
\end{equation}
where $\mathbf{Z}_p^l$ and $\mathbf{Z}_q^l$ are the globally averaged masked visual features, $L$ denotes the total layers of the transformer decoder and ${\rm Sim(\cdot)}$ denotes cosine similarity. We take the calculation of $\mathbf{Z}_p^l$ as an example:
\begin{equation}
\mathbf{Z}_p^l = {\rm Avg}(\hat{\mathbf{M}}_p^l \odot {\rm Downsample}(\hat{\mathbf{Y}}_p)),
\end{equation}
where ${\rm Avg}(\cdot)$ denotes global average pooling, $\odot$ denotes elementwise multiplication, and ${\rm Downsample}(\cdot)$ denotes the downsampling operation to adjust the mask to the same size as the visual feature. 

\begin{figure}[!htbp]
    \centering
    \includegraphics[width=0.8\linewidth]{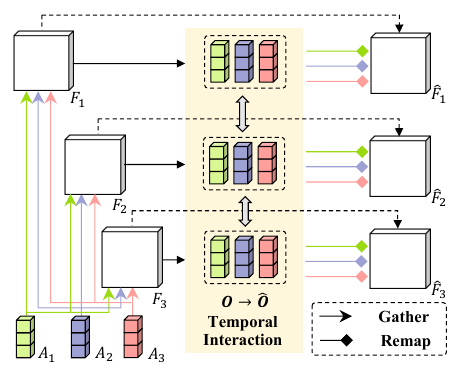}
    \caption{Illustration of ABTI module. Audio features are first leveraged to gather audio-relevant object features from different frames. After exchanging information between these object features, the updated object features are then remapped and added to the original features for enhancement.}
    \label{fig:abti}
\end{figure}
\subsection{Audio Bridged Temporal Interaction}
\label{sec:abti}
The proposed ABTI module aims to endow the frame-level features with temporal context information without introducing too much unnecessary background interaction.
Thus, audio feature is exploited to bridge the interactions among different frames by extracting audio-relevant object features from different frames and exchanging information between these object features, which are then remapped back for feature enhancement.
ABTI is inserted after each scale of visual features and we take one of them as an example for detailed elaboration, which is denoted as $\mathbf{F}\in \mathbb{R}^{T\times H \times W \times C_v}$ in this section only.
$H$ and $W$ represent the height and width of $\mathbf{F}$.
As illustrated in Figure~\ref{fig:abti}, there are three steps in the ABTI module, including object feature gathering, temporal interaction, and object feature remapping.
First, for the sounding objects associated with the $p$-th audio feature, we extract its compact feature at the $q$-th frame from $\mathbf{F}_q$ by feature similarity as follows:

\vspace{-0.45cm}
\begin{gather}
\mathbf{S}_{pq} = {\rm Softmax}(f_a(\mathbf{A}_p)\otimes f_m(\mathbf{F}_q)^T), \\
\mathbf{O}_{pq} = \mathbf{S}_{pq} \otimes f_n(\mathbf{F}_q),
\end{gather}
where $\mathbf{A}_p$ and $\mathbf{F}_q$ are first reshaped to $\mathbb{R}^{1\times C_v}$ and $\mathbb{R}^{HW\times C_a}$ respectively, $f_a(\cdot)$, $f_m(\cdot)$, and $f_n(\cdot)$  change the channel numbers of input features to $C_b$ by linear transformation. 
Each element of $\mathbf{S}_{pq} \in \mathbb{R}^{1\times HW}$ represents the contribution of the corresponding pixel to the sounding objects, and the compact feature $\mathbf{O}_{pq}\in \mathbb{R}^{C_b}$ is generated through necessary reshaping operation.
We arrange the compact features of all the sounding objects at all the frames as a sequence $\mathbf{O}=\{\mathbf{O}_{pq}\}, p\in [1, T], q\in [1, T]$.
Second, self-attention is applied to $\mathbf{O}$ for temporal interaction and the updated object features are represented as $\hat{\mathbf{O}}=\{\hat{\mathbf{O}}_{pq}\}, p\in [1, T], q\in [1, T]$.
Last, to enhance the original feature $\mathbf{F}_q$ with temporal information from other frames, the updated object features are remapped back to match the shape of $\mathbf{F}_q$ as follows:
\begin{equation}
\tilde{\mathbf{F}}_{pq} = \mathbf{S}_{pq}^T\otimes f_o(\hat{\mathbf{O}}_{pq}),
\end{equation}
where $\hat{\mathbf{O}}_{pq}$ is first reshaped to $\mathbb{R}^{1\times C_b}$ and then transformed by $f_o(\cdot)$ to the channel number of $C_f$.
After change the shape of $\tilde{\mathbf{F}}_{pq}$ to $\mathbb{R}^{H\times W\times C_v}$, the enhanced feature $\hat{\mathbf{F}}_q$ is generated by as follows:
\begin{equation}
\hat{\mathbf{F}}_q = \mathbf{F}_q + \sum_{p=1}^{T}{\tilde{\mathbf{F}}_{pq}}.
\end{equation}
The above operations are equally performed on all the visual features to enable cross-frame interaction.

\subsection{Loss Functions}
The overall loss function is formulated as follows:
\begin{equation}
\mathcal{L} = \mathcal{L}_{mask} + \lambda_{sim}\mathcal{L}_{sim},
\end{equation}
where $\lambda_{mask}$ consists of Binary Cross-Entropy Loss and Dice Loss~\cite{diceloss} following the implementation of~\cite{Mask2Former}:
\begin{equation}
\lambda_{mask} = \lambda_{bce}\mathcal{L}_{bce} + \lambda_{dice}\mathcal{L}_{dice},
\end{equation}
the $\mathcal{L}_{bce}$ and $\mathcal{L}_{dice}$ are equally applied between the output of each decoder layer and the ground-truth mask as supervision.

\section{Experiments}
\subsection{Datasets and Evaluation Metrics}
We conduct experiments on two benchmark settings of the AVS~\cite{AVS}: 1) The semi-supervised Single Sound Source Segmentation (S4) where the sounding object remains the same in the given video clip, i.e., only the mask annotation of the first frame is provided for training; 2) The fully supervised Multiple Sound Source Segmentation (MS3) where the sounding object dynamically changes over time, and mask annotations of all the $T$ frames are provided. 
F-score $M_\mathcal{F}$ and Jaccard index $M_\mathcal{J}$ are adopted as evaluation metrics. 
$M_\mathcal{F}$ calculates the harmonic mean of pixel-level precision and recall, while $M_\mathcal{J}$ calculates the intersection-over-union between the predicted mask and the ground truth mask.

\subsection{Implementation Details}
We use both ResNet-50~\cite{resnet} pretrained on MSCOCO~\cite{coco} dataset and the PVT-v2 b5~\cite{pvt} pretrained on ImageNet~\cite{imagenet} dataset as the visual encoders. 
The total number of video frames $T$ is set to $5$ for each video clip.
The number $N$ of transformer decoder stage is set $3$.
$\lambda_{bce}$, $\lambda_{dice}$ and $\lambda_{sim}$ are all set to $1$.
For the S4 setting, we use the polynomial learning rate schedule and the AdamW optimizer with an initial learning rate of $1.25e^{-4}$ and weight decay of $5e^{-2}$. 
Batch size is set to 8/4 and the total number of training iterations is set to 20k/40k for experiments on ResNet-50/PVT-v2 b5.
The MS3 setting adopts the same hyperparameters except: (1) the initial learning rate is $5e^{-4}$, (2) the total training iterations are reduced to 2k/4k.
More implementation details are provided in the supplementary materials.

\begin{table*}[!htb]
\tabcolsep=0.3cm
\renewcommand\arraystretch{1.1}
\begin{center}
\begin{tabular}{c||c|c|cc|cc|cc}
\hline \thickhline
\rowcolor{mygray}
 & &  & \multicolumn{2}{c|}{S4} & \multicolumn{2}{c|}{MS3} & \multicolumn{2}{c}{MS3$^*$} \\ 
\rowcolor{mygray}
\multirow{-2}*{Method} & \multirow{-2}*{Task} & \multirow{-2}*{Backbone} & $M_\mathcal{F}$ & $M_\mathcal{J}$ & $M_\mathcal{F}$ & $M_\mathcal{J}$ & $M_\mathcal{F}$ & $M_\mathcal{J}$ \\ 
\hline \hline 
MSSL~\cite{qian2020multiple} & SSL & ResNet-18 & 0.663 & 0.449 & 0.363 & 0.261 & - & - \\ 
\hline
SST~\cite{duke2021sstvos} & VOS & ResNet-101 & 0.801 & 0.663 & 0.572 & 0.426 & - & - \\ 
\hline \hline 
\multirow{2}*{TPAVI~\cite{AVS}} & \multirow{2}*{AVS} & ResNet-50 & 0.848 & 0.728 & 0.578 & 0.479 & - & 0.543 \\
 & & PVT-v2 & 0.879 & 0.787 & 0.645 & 0.540 & - & 0.573 \\ 
\hline
\multirow{2}*{AQFormer(ours)} & \multirow{2}*{AVS} & ResNet-50 & \textbf{0.864} & \textbf{0.770} & \textbf{0.669} & \textbf{0.557} & \textbf{0.699} & \textbf{0.593} \\
 & & PVT-v2 & \textbf{0.894} & \textbf{0.816} & \textbf{0.721} & \textbf{0.611} & \textbf{0.727} & \textbf{0.622}\\
\hline
\end{tabular}
\caption{\textbf{Comparison with state-of-the-art methods.} Results of both S4 and MS3 settings are reported. MS3$^*$ indicates that the model was first trained on the S4 training data.}
\label{tab:sota}
\end{center}
\end{table*}

\begin{table*}
    \tabcolsep=0.275cm
	\renewcommand\arraystretch{1.1}
    \centering
    \begin{subtable}[t]{0.495\linewidth}
        \centering
        \begin{tabular}{c||cc}
            \hline \thickhline
            \rowcolor{mygray}
            Method & $M_\mathcal{F}$ & $M_\mathcal{J}$ \\ 
            \hline \hline
            Vanilla Object Query & 0.534 & 0.450 \\
            Audio Query$\dagger$ & 0.643 & 0.526 \\
            \textbf{Audio Query} & \textbf{0.647} & \textbf{0.535} \\   
            \hline
        \end{tabular}    
        \caption{\textbf{Effect of audio query}. Audio Query$\dagger$ means each audio query only gathers object information from a single frame.}
        \label{tab:effect}
    \end{subtable}
    \begin{subtable}[t]{0.495\linewidth}
        \centering
        \begin{tabular}{cc||cc}
        \hline \thickhline
        \rowcolor{mygray}
        \rowcolor{mygray}
        ABTI & AVSim Loss & $M_\mathcal{F}$ & $M_\mathcal{J}$ \\ 
        \hline \hline 
         & & 0.647 & 0.535\\
        \checkmark & & 0.656 & 0.543 \\
         & \checkmark & 0.650 & 0.546 \\
        \checkmark & \checkmark & \textbf{0.669} & \textbf{0.557} \\
        \hline
        \end{tabular}
        \caption{\textbf{Component analysis}.}
        \label{tab:comp}
    \end{subtable}
    \begin{subtable}[t]{0.495\linewidth}
        \centering
        \begin{tabular}{c||cccc}
        \hline \thickhline
        \rowcolor{mygray}
        \rowcolor{mygray}
        Inserting Positions & - & 1,2,3 & 4 & 1,2,3,4 \\
        \hline \hline
        $M_\mathcal{F}$ & 0.650 & 0.652 & 0.652 & \textbf{0.669} \\
         $M_\mathcal{J}$ & 0.546 & 0.549 & 0.554 & \textbf{0.557} \\
        \bottomrule     
        \end{tabular}
        \caption{\textbf{Inserting positions of ABTI.}}
        \label{tab:position}
    \end{subtable}
    \begin{subtable}[t]{0.495\linewidth}
        \centering
        \begin{tabular}{c||ccc}
        \hline \thickhline
        \rowcolor{mygray}
        \rowcolor{mygray}
        Number of Stages & 1 & 2 & 3 \\
        \hline \hline
         $M_\mathcal{F}$ & 0.599 & 0.632 & \textbf{0.669} \\
        $M_\mathcal{J}$ & 0.500 & 0.520 & \textbf{0.557} \\ 
        \hline
        \end{tabular}
        \caption{\textbf{Number of transformer decoder stages.} }
        \label{tab:layer}
    \end{subtable}
\caption{\textbf{Ablation Studies.} Models are trained on training set of MS3 and evaluated on its testing set with ResNet-50 as the visual encoder.}
\end{table*}

\subsection{Comparison with State-of-the-art Methods}
Since AVS is a pretty new task and there is only one AVS method TPAVI~\cite{AVS} to compare at present, we follow~\cite{AVS} to adapt methods of related domains to AVS and present the results in Table~\ref{tab:sota}.
Taking ResNet-50 as the visual encoder, our AQFormer outperforms TPAVI by margins of $1.6\%$ $M_\mathcal{F}$ and $4.2\%$ $M_\mathcal{J}$ for the S4 setting. 
For the more challenging MS3 setting, AQFormer achieves $9.1\%$ $M_\mathcal{F}$ and $7.8\%$ $M_\mathcal{J}$ absolute gains over TPAVI, which is a notable improvement. 
The results indicate that the explicit object-level audio-visual correspondence construction leads to more robust performance when multiple objects make sounds at the same time or the sounding objects transfer.
When using a stronger visual encoder like PVT-v2 b5, further performance gains can be observed for both settings.
We also conduct experiments on MS3 by using the model pretrained on S4 for initialization, which also has improvements compared with training from scratch.

\subsection{Ablation Studies}
We conduct all the ablation studies on MS3 benchmark with ResNet-50 as the visual encoder to evaluate the effectiveness of our method.
More results are presented in the supplementary materials.

\textbf{Effect of audio query}.
To verify the superiority of conditioning object queries on audio features over vanilla object queries, we replace audio queries with randomly initialized object queries in the transformer decoder and keep other settings the same for comparison, which is represented as `Vanilla Object Query' in Table~\ref{tab:effect}.
The large performance gap between vanilla object query and audio query indicates that gathering audio-relevant object information by audio queries can better identify the sounding objects for AVS.
Besides, we also revise the cross attention in decoder layers so that each audio query only gathers object information from its corresponding frame rather than the whole video, which is denoted by Audio Query$\dagger$.
Performance drops can be observed on both $M_\mathcal{F}$ and $M_\mathcal{J}$ due to the absence of temporal information, indicating the effectiveness of temporal context gathering.

\textbf{Component Analysis}. 
We conduct ablation studies to explore the impact of ABTI module and the auxiliary loss, which is denoted as `AVSim Loss' in Table~\ref{tab:comp}. 
The first row denotes the basic implementation of our AQFormer, which has already outperformed the previous state-of-the-art by large margins($M_\mathcal{F}+6.9\%$, $M_\mathcal{J}+5.6\%$), which shows the superiority of our new architecture.
As shown in the 2nd and 3rd rows, both ABTI module and AVSim loss bring improvement over the baseline, i.e., $M_\mathcal{F}+0.9\%$ and $M_\mathcal{J}+0.8\%$ for ABTI, and $M_\mathcal{F}+0.3\%$ and $M_\mathcal{J}+1.1\%$ for AVSim loss.
The performance improvement indicates both temporal interaction and audio-visual feature alignment are necessary for AVS task.
When combining the ABTI module and AVSim loss together in the 4th row, we can observe an obvious performance boost in all metrics compared with our baseline.

\begin{figure*}[!t]
    \centering
    \includegraphics[width=0.9\linewidth]{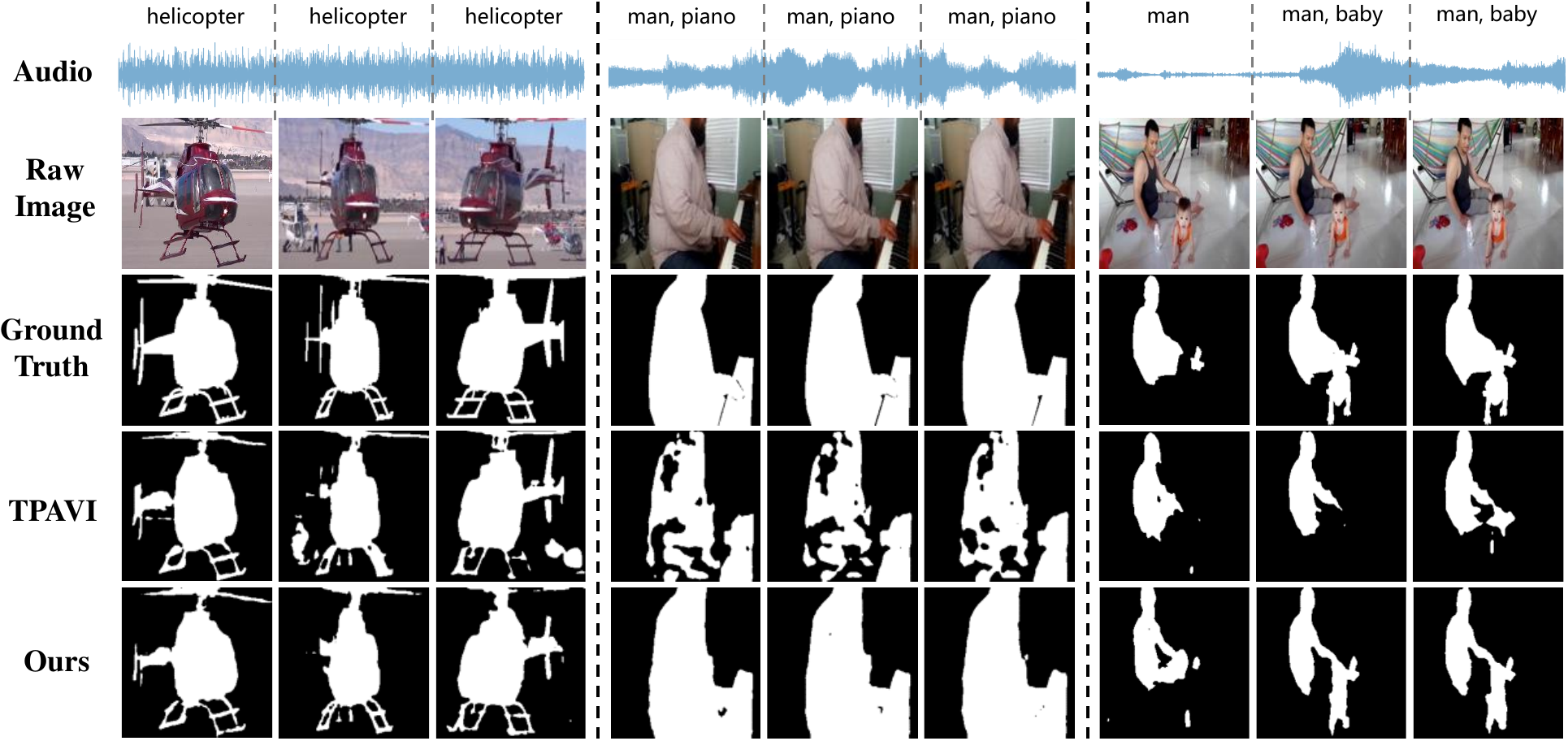}
    \caption{Qualitative comparison with previous method. For each video example, we sample 3 different frames for visualization and arrange them from left to right.}
    \label{fig:mask}
\end{figure*}

\begin{figure}[!t]
    \centering
    \includegraphics[width=0.95\linewidth]{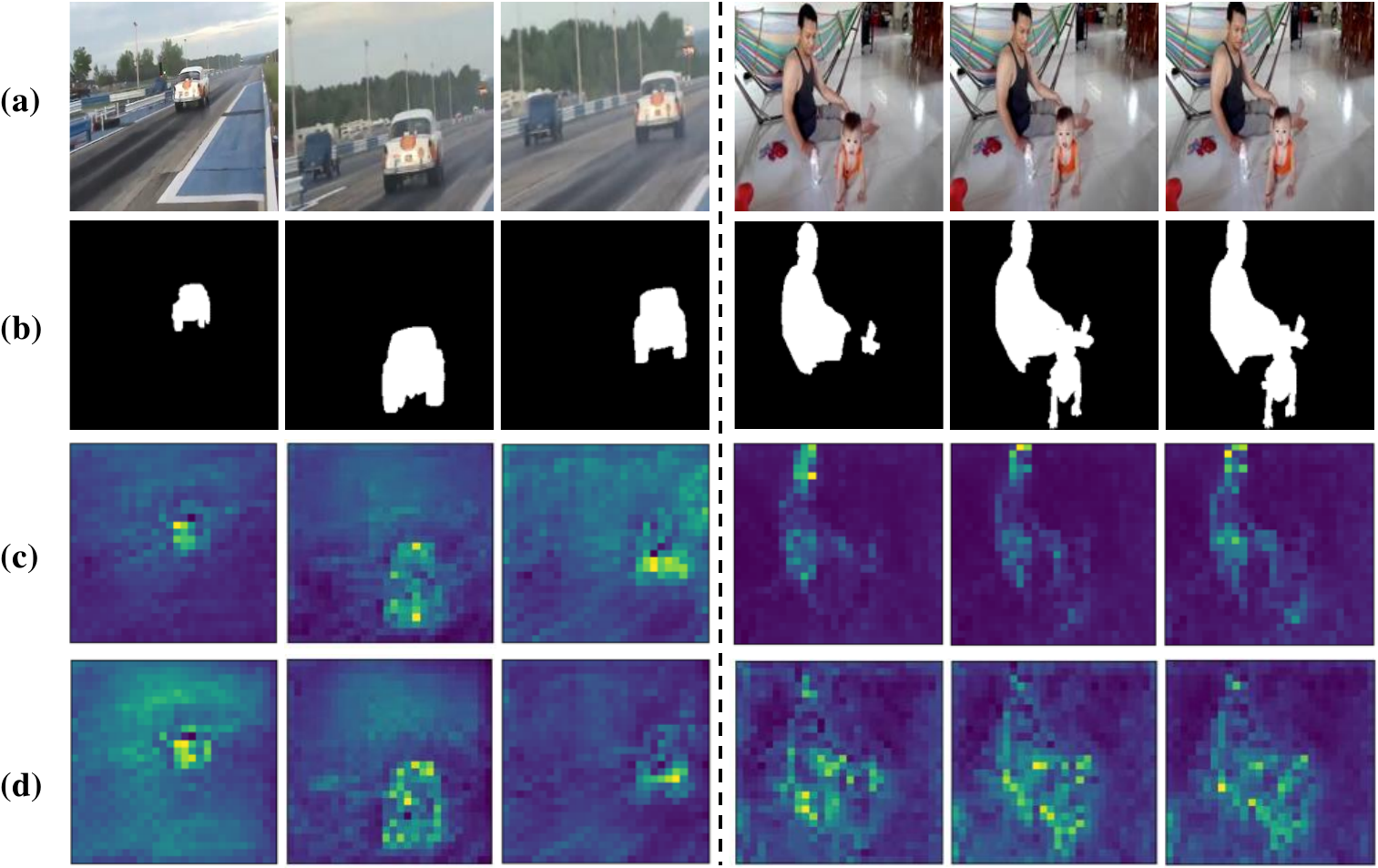}
    \caption{Visualization of attention maps. (a) Raw images of different frames. (b) Ground truth masks. (c) Attention maps of different frames attended by the first audio query. (d) Attention maps of different frames attended by the last audio query. Yellow denotes a larger attention weight. \textit{Better viewed in color}.}
    \label{fig:att}
\end{figure}

\textbf{Inserting Positions of ABTI}.
We evaluate different inserting positions of the ABTI module in the pixel decoder and report the experimental results in Table~\ref{tab:position}. 
It is found that larger gains on $M_\mathcal{J}$ are obtained by applying ABTI to the 4th visual feature that serves as the mask feature for prediction.
Applying ABTI to the first $3$ scales of visual features achieves relatively slight improvement, which may be because that temporal information has already been modeled in the cross-attention steps of the transformer decoder.
The best results are achieved when ABTI is added to all four stages, showing that temporal interaction still helps to get better performances.

\textbf{Number of Transformer Decoder Layers}.
We also adjust the number of transformer stages $N$ and the results are presented in Table~\ref{tab:layer}.
By increasing the number of transformer stages from $1$ to $3$, we can observe consistent performance improvement.
We get the best results when $N$ is set to $3$, $0.669$ $M_\mathcal{F}$ and $0.557$ $M_\mathcal{J}$, which indicates that deeper transformer decoder results in better performances.

\subsection{Qualitative Analysis}
We compare the qualitative results between TPAVI~\cite{AVS} and our AQFormer in Figure~\ref{fig:mask}. 
PVT-v2 is used as the visual encoder of both methods.
Our AQFormer makes better identification and produces finer segmentation results as well.
As shown in the 2nd and 3rd columns of the 1st example, when there are more helicopters appearing in the image but not making any sound, TPAVI cannot figure out the correct sounding objects and includes all of them. 
However, our AQFormer can accurately track the movement of the sounding helicopter without generating false positive predictions.
In the 3rd example, when the sounding objects increase from man only to man and baby, our AQFormer successfully captures the change while TPAVI does not include the baby as the sounding object in the last two frames, which further proves that audio query establishes more robust semantic correspondence between audio and visual signals.

We also visualize the attention maps between the audio queries and the visual features of the last layer of the transformer decoder. 
As illustrated in Figure~\ref{fig:att}, for each audio query, its attention maps at different frames always focus on the regions of the same object(s), even though its associated objects do not make sounds in some frames.
Take the 4th row of the 2nd video as an example, the corresponding audio query is associated with the man and the baby, and only the man makes sound in the first frame.
However, the activated pixels mainly focus on both the man and the baby in the first frame.
These results show that our AQFormer indeed establishes an object-level correlation between audio and video modalities.

\section{Conclusion}
In this paper, we focus on the task of audio visual segmentation (AVS) and propose a new multi-modal transformer framework named Audio-Queried Transformer (AQFormer).
Audio features are utilized to generate conditional object queries to explicitly gather information of the sounding objects from visual features.
An Audio-Bridged Temporal Interaction module is also proposed to interact among multiple frames with the bridging of audio information. 
Extensive experiments on two AVS benchmarks show that our method outperforms the previous method by large margins.
In the future, we hope to explore audio separation techniques and construct more specific instance-level correspondence between audio and visual information.

\section*{Acknowledgements}
This research was supported by National Key R\&D Program of China (2022ZD0115502), National Natural Science Foundation of China (No. 62122010), Zhejiang Provincial Natural Science Foundation of China under Grant No. LDT23F02022F02, and Key Research and Development Program of Zhejiang Province (No. 2022C01082).

\section*{Contribution Statement}
\textbf{Shaofei Huang}$^\dagger$: Designing the architecture of AQFormer, implementing it with code, and writing the paper.

\noindent \textbf{Han Li}$^\dagger$: Designing the architecture of ABTI module,  implementing it with code, and writing the paper.

\noindent \textbf{Yuqing Wang}: Running ablation experiments of AQFormer and writing the experiment part.

\noindent \textbf{Hongji Zhu}: Running some of the ablation experiments of AQFormer and the ABTI module.

\noindent \textbf{Jiao Dai}$^*$: Participating in the discussions regularly, providing guidance, and writing the introduction part.

\noindent \textbf{Jizhong Han}: Participating in the discussions regularly and writing the method part.

\noindent \textbf{Wenge Rong}: Participating in the discussions regularly and writing the related work part.

\noindent \textbf{Si Liu}: Participating in the discussions regularly and writing the abstract part.

\noindent $\dagger$ means equal contributions and $*$ means corresponding author.

\bibliographystyle{named}
\bibliography{ref}

\end{document}